\documentclass[conference]{IEEEtran}
\IEEEoverridecommandlockouts
\IEEEpubid{\makebox[\columnwidth]{979-8-3503-1579-0/24\$31.00 \copyright 2024 IEEE \hfill} \hspace{\columnsep}\makebox[\columnwidth]{}}
\usepackage{cite}
\usepackage{float}
\usepackage{graphicx} 
\usepackage{amsmath,amssymb,amsfonts}
\usepackage{algorithmic}
\usepackage{graphicx}
\usepackage{textcomp}
\usepackage{xcolor}
\usepackage{multirow}
\def\BibTeX{{\rm B\kern-.05em{\sc i\kern-.025em b}\kern-.08em
    T\kern-.1667em\lower.7ex\hbox{E}\kern-.125emX}}

\usepackage{caption,subcaption}
\begin{document}

\title{CBGT-Net: A Neuromimetic Architecture for Robust Classification of Streaming Data\\
\thanks{This work was supported by the Army Research Lab awards W911NF-19-2-0146 and W911NF-22-2-0115, and AFOSR / AFRL award FA9550-18-1-0251. \\
\textsuperscript{*} These authors contributed equally to this work. \\
\copyright 2024 IEEE.  Personal use of this material is permitted.  Permission from IEEE must be obtained for all other uses, in any current or future media, including reprinting/republishing this material for advertising or promotional purposes, creating new collective works, for resale or redistribution to servers or lists, or reuse of any copyrighted component of this work in other works.
}
}

\author{\IEEEauthorblockN{Shreya Sharma\textsuperscript{*}}
\IEEEauthorblockA{\textit{Robotics Institute} \\
\textit{Carnegie Mellon University}\\
Pittsburgh, PA, United States \\
ssharma5@andrew.cmu.edu}
\and
\IEEEauthorblockN{Dana Hughes\textsuperscript{*}}
\IEEEauthorblockA{\textit{Robotics Institute} \\
\textit{Carnegie Mellon University}\\
Pittsburgh, PA, United States \\
danahugh@andrew.cmu.edu}
\and
\IEEEauthorblockN{Katia Sycara}
\IEEEauthorblockA{\textit{Robotics Institute} \\
\textit{Carnegie Mellon University}\\
Pittsburgh, PA, United States \\
sycara@andrew.cmu.edu}
}

\maketitle
\IEEEpubidadjcol

\begin{abstract}
This paper describes CBGT-Net, a neural network model inspired by the cortico-basal ganglia-thalamic (CBGT) circuits found in mammalian brains.  Unlike traditional neural network models, which either generate an output for each provided input, or an output after a fixed sequence of inputs, the CBGT-Net learns to produce an output after a sufficient criteria for evidence is achieved from a stream of observed data.  For each observation, the CBGT-Net generates a vector that explicitly represents the amount of evidence the observation provides for each potential decision, accumulates the evidence over time, and generates a decision when the accumulated evidence exceeds a pre-defined threshold.  We evaluate the proposed model on two image classification tasks, where models need to predict image categories based on a stream of small patches extracted from the image.  We show that the CBGT-Net provides improved accuracy and robustness compared to models trained to classify from a single patch, and models leveraging an LSTM layer to classify from a fixed sequence length of patches.
\end{abstract}



\section{Introduction}

Significant strides in deep learning have led to remarkable advancements across various domains such as image classification, natural language comprehension, and decision-making
~\cite{pouyanfar2018-survey}.  The success of such methods arises from multi-layered architectures capable of learning feature mappings at increasing levels of abstraction from large datasets.  Despite the success of deep learning, models are trained in an end-to-end manner and generally produce an output for every provided input. 
Generated output may be incorrect---often with a high level of confidence---with minimal perturbation to the input~\cite{su2019-one_pixel}, and traditional neural network models are not designed to consider when a single input is insufficient for inference purposes.  For instance, traditional image classification models generate a category for a single image without the ability to consider additional viewpoints, while policies learned for control are designed to generate an action regardless of how complete or noisy the observation is.

In contrast, models of decision-making in primate brains have been developed where decisions are made based on the integration of noisy information over time~\cite{smith2004_decisions}.  In these models, evidence for a response is accumulated until a requisite amount is reached, 
explaining response accuracy and timing.  Specifically, the cortico-basal ganglia-thalamic (CBGT) circuits in the brain have been shown to play a role in action selection~\cite{maia11_basal_ganglia,stocco2010_basal_ganglia}, including describing means of evidence accumulation for and response criteria of competing actions~\cite{pisauro2017_neural}.  In essence, this 
circuitry deliberates over potential actions based on a stream of noisy or incomplete information from multiple cortical areas. 

Inspired by the evidence accumulation aspect of primate decision-making, our research aims to develop and evaluate a neuromimetic model of the CBGT circuit in mammalian brains.  
We believe such a model would provide several desired features in autonomous decision-making---in addition to potentially improved model accuracy, 
the deliberation process of the model is transparent, allowing for better interpretability during human-autonomy collaborations.  Building on our prior proof-of-concept work in this area~\cite{agarwal2018-cbgt}, we present a CBGT-inspired neural network architecture\footnote{Code available at https://github.com/ShreyaSharma99/CBGT-Net} and evaluate its ability to learn to integrate noisy information, as well as determine the effect of varying evidence criterion, in complex domains (i.e., vision-based tasks).  We demonstrate that the proposed model is able to perform classification tasks using a stream of incomplete information more accurately than models trained to classify based on a single observation, and also generally outperforms LSTM-based sequential models in terms of accuracy and data efficiency.  Additionally, our model's performance is robust to decreasing information in observations, compared to the LSTM models.  Finally, our model is designed to make decisions based on acquiring a sufficient amount of evidence, as opposed to a fixed amount of time, which is easily adjusted during deployment using a simple decision threshold level.

This paper is organized as follows:  Section~\ref{sec:related_work} describes relevant work related to our approach; Section~\ref{sec:cbgt} provides a brief 
description of the 
CBGT circuit in mammalian brains; Section~\ref{sec:approach} describes the architecture and training approach for our model; Section~\ref{sec:evaluation} and~\ref{sec:results} describes our evaluation 
and results.  Section~\ref{sec:future_work} provides discussion and future work. 


\section{Related Work}\label{sec:related_work}

In the field of neuroscience, computational models of basal ganglia circuitry have been used to explore aspects of decision making in dynamic environments and its role in reinforcement learning.  For instance, competition between neural pathways in the basal ganglia has been proposed as a model of action uncertainty~\cite{dunovan2016-believer} and for describing exploration-exploitation tradeoffs in volatile environments~\cite{bond2021-dynamic}. While previous research has showcased the basal ganglia's involvement in different facets of decision-making, the existing models predominantly investigate 
the biological dimensions of decision-making, such as response time. In contrast, our focus lies in developing models tailored for machine learning tasks with inspiration drawn from neuroscience.


In the area of deep neural networks, confidence-aware learning aims to not only accurately perform some inference task (e.g., image classification), but to also assign a confidence score to each inference.  In~\cite{moon2020-confidence}, a correctness ranking loss is utilized to ordinally rank training examples and produce a confidence score for classification tasks.  In~\cite{mandelbaum2017-distance}, training loss is augmented with a distance loss to encourage clustering of training examples in an embedding space; post-training, the distance of a novel data point to the nearest neighbor of the training data in the embedding space is leveraged as a confidence score.  While such confidence scores are analogous to our usage of evidence, these approaches aim to generate confidence scores of a single prediction, while in our approach, the evidence encoder learns to produce a value akin to confidence when learning with a stream of data.

Our prior exploration into developing a CBGT-inspired network~\cite{agarwal2018-cbgt} demonstrates a proof-of-concept network capable of learning decision thresholds and very simple encoders; in this paper, we extend this effort to learn encoders for more complex data streams (e.g., images), demonstrate that our approach is agnostic to encoding layers, and utilize a more effective supervised training approach.


\section{Cortico-Basal Ganglia Thalamic Circuit}\label{sec:cbgt}

The network architecture presented in this paper is inspired by CBGT circuits in mammalian brains, and the role they play in decision making and evidence accumulation~\cite{maia11_basal_ganglia,stocco2010_basal_ganglia,pisauro2017_neural}.  
Corticostriatal connections provide pathways for projection from the functional areas of the cortex---including the sensorimotor, associative, and limbic areas---to the striatum.  For each potential action (i.e., motor neuron activation), two pathways exist in the basal ganglia that facilitate or suppress the action in the thalamus:  Direct (``Go'') pathways inhibit the globus pallidus internus (GPi), which in turn causes disinhibition of the thalamus and facilitation of the action corresponding to the circuit; Indirect (``NoGo'') pathways inhibit the globus pallidus externus (GPe), which in turn disinhibits the GPi and suppresses the circuit's action.

In the context of the described structure, the information used for decision-making is generated in the cortex, which in turn increases or decreases the total activation of the ``Go'' and ``NoGo'' pathways for each action.  Action selection for a given action is based on the relative activation of the ``Go'' and ``NoGo'' pathways---actions with a higher differential between the ``Go'' and ``NoGo'' pathways are more likely to be performed.  In essence, the basal ganglia facilitates actions with a probability proportional to the activation difference between the ``Go'' and ``NoGo'' pathways.  An action is performed once the activation difference for the action exceeds some criteria.  Tonic dopamine levels in the brain increase the excitability of the ``Go'' pathways and decrease the excitability of the ``NoGo'' pathways, which influences the overall criteria for action selection, as well as reaction time.


\section{Technical Approach}\label{sec:approach}

\subsection{Network Architecture}\label{sec:network_arch}




In this section, we describe a neural network model, referred to as \textit{CBGT-Net}, whose functionality aims to be analogous to the functionality of CBGT circuits in mammalian brains, as described in Section~\ref{sec:cbgt}.  Unlike traditional feed-forward networks, this model is designed to perform inference tasks based on a \textit{stream} of observations, as opposed to a single input.  In contrast to recurrent architectures, which maintain arbitrary latent embeddings as an internal state, the model maintains the total evidence accumulated over time in support of each possible decision. 
We note that the inference task described here differs from inference tasks performed by recurrent neural networks on \textit{sequential} data:  recurrent models generate a decision at a fixed point in time, or at the end of a sequence of known length; the inference task here requires the model to make a decision at an arbitrary point in time, in the presence a (hypothetically) unending stream of data. 


The model accepts as input a stream of observations at discrete time steps, denoted $\mathbf{o}_t$, from an environment (see Section~\ref{sec:environments}).  At each time step, the model produces a pair of outputs: an output vector, $\mathbf{y}_t$, corresponding to the inference task, and a binary decision variable, $d_t$, indicating if the model has accumulated sufficient evidence to make a decision.  
 The output of the model should only be considered meaningful at the first time step that the decision variable indicates that the evidence criteria is satisfied, denoted as $t_d$. Thus, while the model generates a pair of outputs at each time step, only the decision at the first changepoint is considered meaningful.
 
\begin{figure}
    \centering
    \includegraphics[width=0.9\columnwidth]{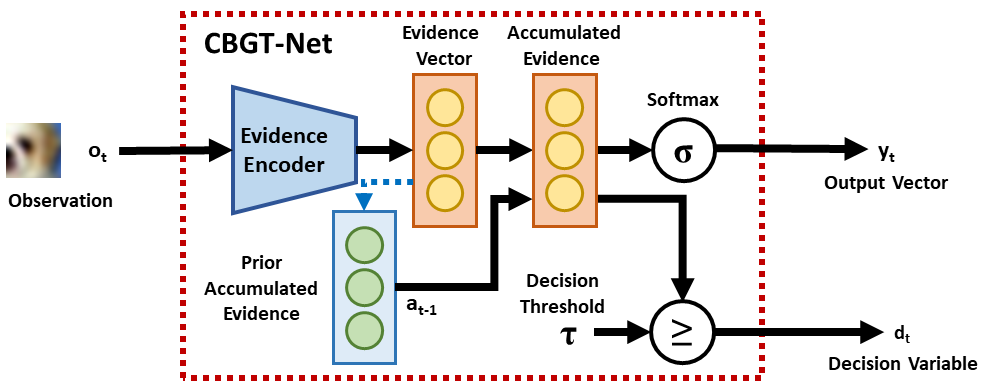}
    \caption{Main components of the CBGT-Net architecture.} 
    \label{fig:cbgt_arch}
\end{figure}

For this paper, we explore the task of classifying observation streams, thus the output vector is interpreted as the probability distribution over possible categories.  Figure~\ref{fig:cbgt_arch} shows the basic structure of the model and the interaction of its core components---Evidence Encoder, Evidence Accumulator, and Decision Threshold Module---each of which is detailed below.

The \textit{Evidence Encoder}, $E_\theta$ is a parameterized model responsible for mapping observations at each time step $t$, called $\mathbf{o}_t$, to an evidence vector, $\mathbf{e}_t$,

\begin{equation}
    \mathbf{e}_t = E_\theta(\mathbf{o}_t)
\end{equation}

where $\theta$ represents the parameters of the evidence encoder.  The evidence encoder may be an arbitrary neural network model suitable for the modality of the observation data (e.g., convolutional neural network for images); 
we constrain its design to
generate an output whose dimensionality and semantic interpretation are consistent with the available decisions. 
For example, for classification tasks, each decision category can have a single corresponding element in the evidence vector $\mathbf{e}_t$. 




The \textit{Evidence Accumulator} consists of a vector corresponding to the total evidence accumulated since the beginning of the input stream, $\mathbf{a_t}$,

\begin{equation}
    \mathbf{a}_t = \mathbf{a}_{t-1} + \mathbf{e}_t
\end{equation}

with $a_0$ assumed to be $\mathbf{0}$.  In addition, the accumulated evidence is mapped to the output vector using a suitable mapping function.  For classification tasks, this simply involves calculating the \textit{softmax} over the accumulated values

\begin{equation}
    \mathbf{y}^{(i)}_t = \frac{exp(\mathbf{a}^{(i)}_t)}{\sum_{j=1}^{K} exp(\mathbf{a}^{(j)}_t)}
\end{equation}

where $(i)$ corresponds to the \textit{i}$^{th}$ element of a vector, and $K$ is the number of decision categories.

The \textit{Decision Threshold} module is a component that is used to determine if the required evidence criterion has been satisfied and generates the decision variable, $d_t$.  For this paper, this module is defined by a fixed threshold parameter, $\tau$.  For each time step, the decision variable is \textit{true} if and only if at least one element in the evidence accumulator exceeds this threshold,

\begin{equation}
  d_t =
  \begin{cases}
      true & \text{if} \hspace{4pt} \exists i \hspace{2pt} \text{where} \hspace{2pt} \mathbf{a}^{(i)}_t \geq \tau \\
      false & \text{otherwise}
  \end{cases}
\end{equation}

If the threshold is not exceeded, the model ingests additional data, allowing for additional evidence before making a choice.  

At the initial instance when $d_t$ becomes \textit{true}, signifying the first time the threshold is crossed, the model makes a prediction by selecting the category associated with the highest value in the Evidence Accumulator's output vector $y_t$.

\section{Evaluation}\label{sec:evaluation}


\subsection{Environments}\label{sec:environments}

To evaluate the described approach, we developed a set of environments to generate streams of information for use as input to the model, where individual streams are conditioned on a target category.  Environments were constructed around publicly available datasets used for image classification.

We denote a single stream of information generated by the environment as an episode.  At the beginning of a given episode, the environment selects an image at random from the dataset and its corresponding target category.  At each time step, the environment extracts a square patch of pixels from the environment at a random location in the image.  The extracted patch is zero-padded to produce an image with the same dimensionality as the original dataset, and in such a manner that the patch is centered on the image.  This approach ensures that the qualitative amount of information present in a single observation in the episode can be controlled (through the size of the patch), and positional information regarding the observation is removed (through centering of the patch).  
The task of the model is to infer the target category of the selected image based on the stream of observations.

\begin{figure}
    \centering
    \includegraphics[width=0.85\columnwidth]{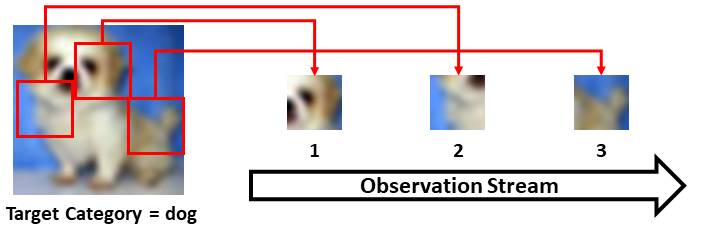}
    \caption{Example episode from CIFAR-10 environment:  a sequence of three patches from an image in the ``dog'' category.}
    \label{fig:environment}
\end{figure}

We constructed environments based on two image datasets:

\subsubsection{MNIST Environment} The MNIST dataset~\cite{lecun2010-mnist,lecun1998-lenet5} consists of images of handwritten digits, with ten target categories corresponding to each digit (i.e., 0 -- 9).  Each image is greyscale and 28x28 pixels in size and contains a single handwritten digit.  The dataset contains a total of 60,000 training images and 10,000 test images. Using this dataset, we constructed environments which generated patches of size 5x5, 8x8, 10x10, 12x12, 16x16, 20x20.

\subsubsection{CIFAR-10 Environment} The CIFAR-10 dataset~\cite{Krizhevsky2009LearningML} consists of 50,000 color images in training data and 10,000 color images for testing.  
Each image is 32x32 pixels in size.  The ten image categories 
are \textit{airplane}, \textit{automobile}, \textit{bird}, \textit{cat}, \textit{deer}, \textit{dog}, \textit{frog}, \textit{horse}, \textit{ship}, and  \textit{truck}.  The environments which were constructed using this dataset generated patches of size 5x5, 8x8, 10x10, 12x12, 16x16, 20x20. 




\subsection{Evidence Encoders}

For evaluation, we utilized existing network architectures as evidence encoders in the CBGT-Net.  For experiments involving the MNIST Environment, we utilized Lenet-5~\cite{lecun1998-lenet5} as the evidence encoding network.  Lenet-5 is a convolutional neural network consisting of seven total layers---two convolutional layers interleaved with two subsampling layers, followed by two fully connected layers and a softmax classification layer.

For experiments involving the CIFAR-10 Environment, we utilized a ResNet style residual architecture~\cite{he2016-resnet}. The model consists of an initial convolutional layer and batch norm layer, followed by six ``blocks'' of two convolutional layers, followed by a fully connected layer, an average pooling layer, and softmax classification layer.  Each block in the model is designed to maintain the size of the generated feature map and includes a shortcut connection from the input of the block to the output so that each block learns to compute a residual, rather than general, mapping from input to output.  Additionally, the network downsamples the size of the feature map after every pair of blocks.




\subsection{Baselines} \label{subsec:baselines}

For each experiment, we compare our approach with multiple baselines.  As with the CBGT-Net, all baseline models are trained to minimize the cross-entropy loss given in Equation~\ref{eq:cross_entropy_loss}.

\subsubsection{Single Patch Evidence Encoder}
For each experiment, we train the evidence encoder used in the CBGT-Net to predict the target category of an image from a single patch.  This baseline serves as a benchmark for evaluating the encoder's capability to independently classify individual data patches. Furthermore, it helps us assess how effectively the model's accuracy improves when evidence is accumulated from multiple observed patches.


\subsubsection{LSTM Model} For each experiment, we train a model in which the output of the evidence encoder is connected to a Long Short Term Memory (LSTM) layer~\cite{hochreiter1997-lstm}.  The LSTM layer has ten memory cells and is provided with a sequence of evidence encoder outputs from observations from the environment.  Models were trained on sequences of varying length, in order to compare model performance with the CBGT-Net's decision times at each decision threshold.
The model outputs the predicted category at the final time step. 

\subsection{Training Details}

For training purposes, we utilize the cross-entropy between the output vector of the CBGT-Net and the target category at the decision time, $t_d$, as the objective function to minimize,

\begin{equation}\label{eq:cross_entropy_loss}
    \mathcal{L}_{CE} = - log(\mathbf{y}^{(T)}_{t_d})
\end{equation}

where $T$ is the index of the target category to be classified.  



For each experiment, models were trained 
using Adam to optimize model parameters~\cite{kingma2014-adam}. Learning rate for the optimizer was set to 1e-3, and a batch size of 512 episodes per training epoch was used for all experiments.

\subsection{Evaluation Measures}
To evaluate our models, we calculate the accuracy, average decision time, and the number of training episodes required. 
Accuracy measures the percentage of correct predictions the model makes when tested with a batch of episodes. Average decision time, on the other hand, quantifies the average number of steps taken 
before the model reaches a decision. In simpler terms, average decision time measures how much input data the model needs to see before making a confident prediction i.e. before it crosses the predefined threshold.

For each model and environment, we calculated the number of training examples required for the model to converge.  During training, validation accuracy was calculated after every two training epochs (i.e., after training with 1,024 episodes).  We performed exponential smoothing on the validation accuracy, with a smoothing factor of $\alpha=0.995$.  The normalized root mean standard deviation (NRMSD, i.e., standard deviation normalized to the mean) of the validation accuracy was calculated at each step using a window over the previous 100 steps.  Training is considered converged when the NRMSD is below an empirically determined threshold of 0.0015.

\section{Results}\label{sec:results}

\begin{figure*}[ht!]
    \centering 
    \includegraphics[width=0.9\linewidth]{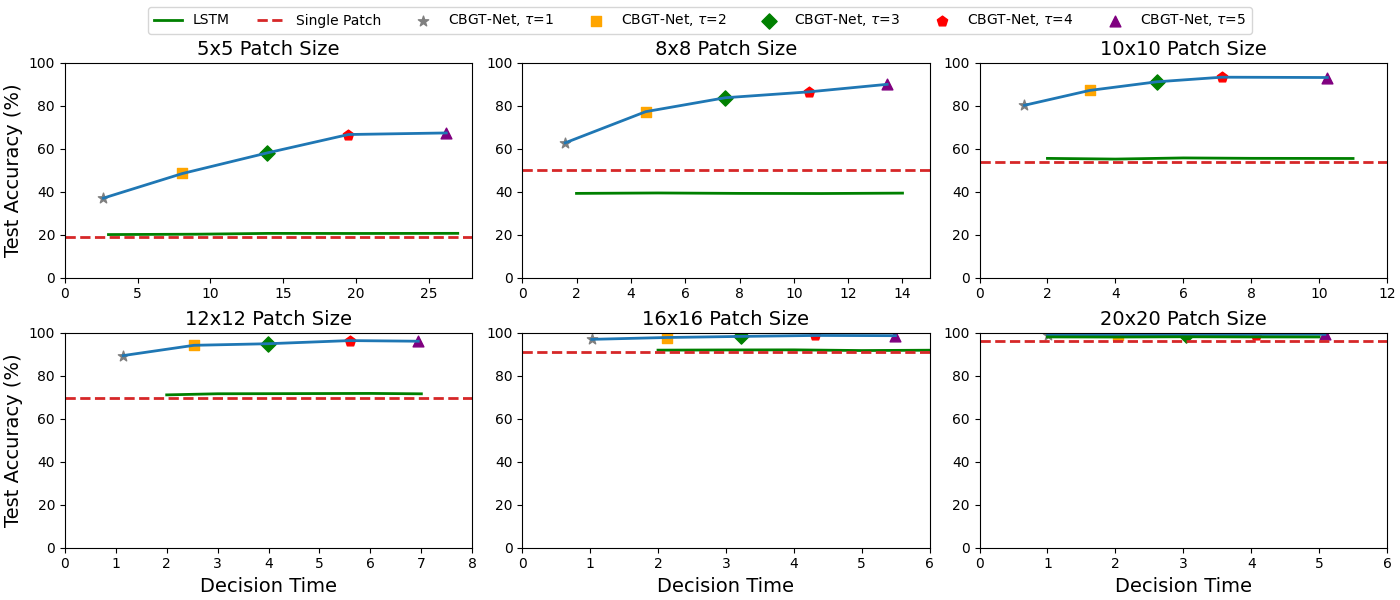}
    \caption{Inference accuracy of the CBGT-Net and baselines as a function of decision time for the MNIST Environment.  Markers on the CBGT-Net results indicate the \textit{average} decision time for models with a given threshold value.  LSTM models were trained with sequence lengths corresponding to the nearest value above corresponding CBGT-Net decision times.}
    \label{fig:mnist_baselines_compare}
\end{figure*}

\begin{figure*}[ht!]
    \centering 
    \includegraphics[width=0.9\linewidth]{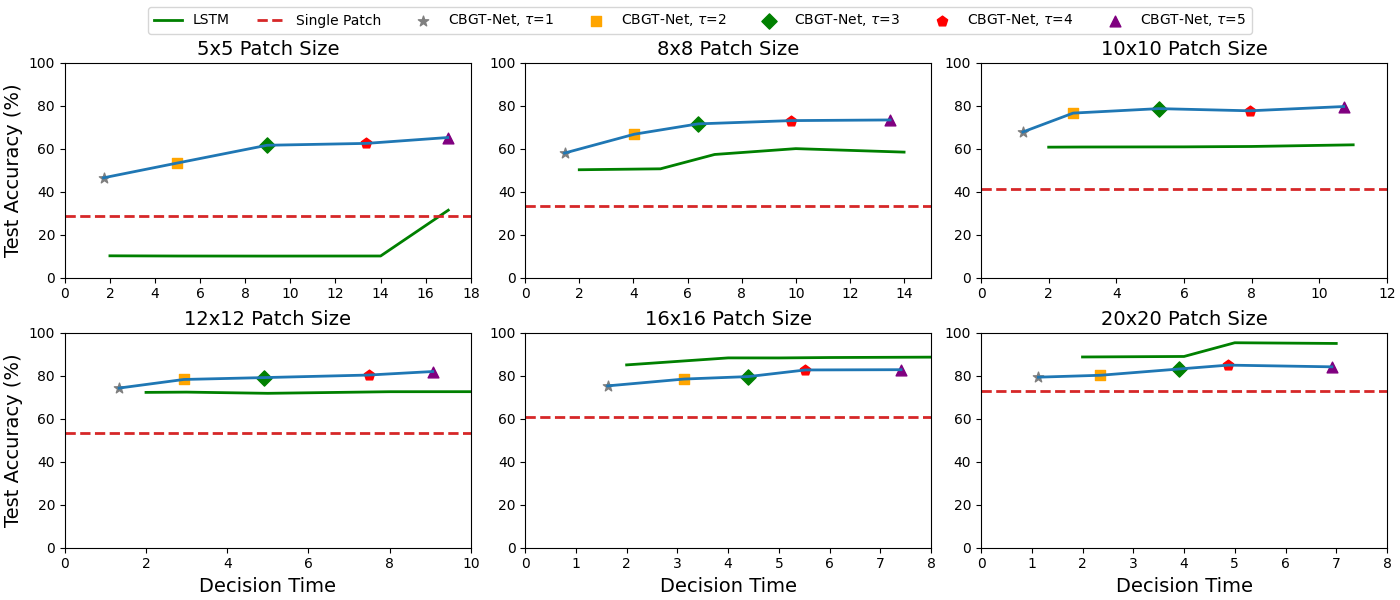}
    \caption{Inference accuracy of the CBGT-Net and baselines as a function of decision time for the CIFAR-10 Environment.  Markers on the CBGT-Net results indicate the \textit{average} decision time for models with a given threshold value.  LSTM models were trained with sequence lengths corresponding to the nearest value above corresponding CBGT-Net decision times.}
    \label{fig:cifar_baselines_compare}
\end{figure*}

Figures~\ref{fig:mnist_baselines_compare} and~\ref{fig:cifar_baselines_compare} compare the performance of the CBGT-Net and baseline models for MNIST 
and CIFAR-10 Environments, respectively.  Each figure compares the inference accuracy of the models based on the amount of information in each observation (i.e., patch size), and the number of observations made.  For the CBGT-Net results, markers indicate the average decision time for decision thresholds in the range of 1 to 5; results for LSTM models were extracted from models trained with a sequence length comparable to the CBGT-Net decision times.  Additionally, the accuracy of evidence encoders trained to categorize a single patch is provided for each case.

In general, the CBGT-Net outperforms both the LSTM and single patch baselines across decision times, with the exception that the LSTM models outperform the CBGT-Net models on the CIFAR-10 Environments with 16x16 and 20x20 patch sizes.  For the MNIST Environments, the LSTM models have roughly the same accuracy as the single patch models, demonstrating that this model was unable to learn to leverage the multiple observations to improve performance for this environment.  The CBGT-Net, on the other hand, not only demonstrates an improvement in performance as sequence length increases, indicating that the model benefits from additional evidences to a certain extent, but also shows significant robustness when each observation's patch size decreases.

For the CIFAR-10 Environments, the LSTM models demonstrate the ability to outperform the single patch baseline, demonstrating its ability to improve its performance with multiple observations (with the notable exception of the 5x5 patch size environment).  For these environments, the CBGT-Net shows improvement over the single patch models similar to the MNIST environments; the performance margin between the CBGT-Net and LSTM models for environments using smaller patch sizes also demonstrates the CBGT-Net's improved robustness to reduced information in each observation when compared to both the LSTM and single patch baselines.

\begin{figure}
  \centering
  \begin{subfigure}{0.85\columnwidth}
    \includegraphics[width=\linewidth]{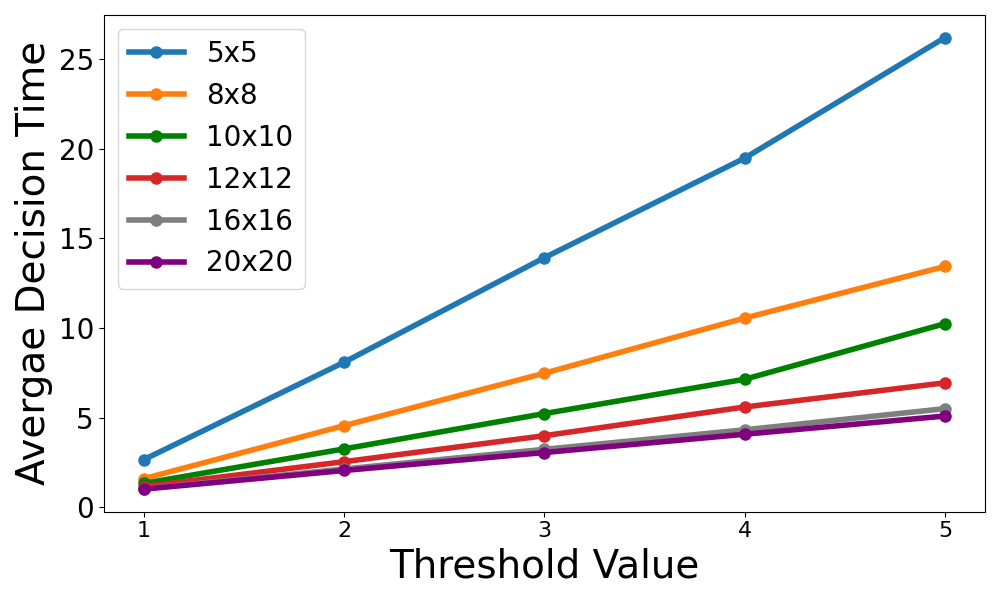}
    \caption{MNIST Environments}
    \label{fig:subplot1}
  \end{subfigure}


  \begin{subfigure}{0.85\columnwidth}
    \includegraphics[width=\linewidth]{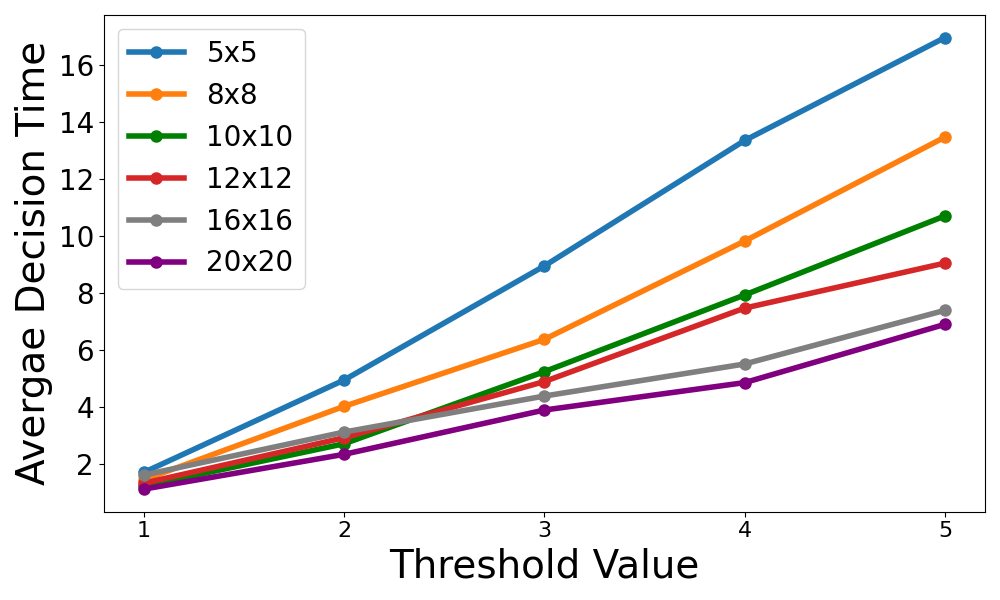}
    \caption{CIFAR10 Environments}
    \label{fig:subplot2}
  \end{subfigure}

  \caption{Average Decision Time taken by CBGT\_Net trained at different threshold values ($\tau$) on MNIST (Fig. a) and CIFAR10 (Fig. b) Environments for different patch size observations}
  \label{fig:threshVsDT}
\end{figure}

Figure ~\ref{fig:threshVsDT} shows the average decision time for the CBGT-Net for different decision thresholds and patch sizes for the MNIST Environments and CIFAR-10 Environments.  As can be seen, the required decision time increases as either decision threshold increases or patch size decreases.  For larger patch sizes in the MNIST Environments (i.e., 16x16 and 20x20), the decision time is roughly equivalent to the decision threshold---this relationship indicates that the generated evidence vector is, on average, producing a maximum value (i.e., 1) for the target category, and that a single patch at these sizes is likely sufficient for categorization purposes.




\begin{figure}[ht!]
    \centering
    \includegraphics[width=0.95\linewidth]{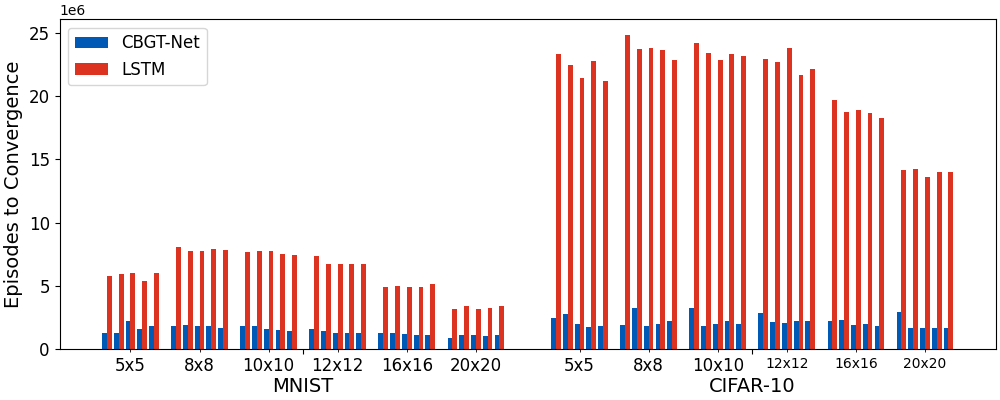}
    \caption{Number of training episodes required for convergence of CBGT-Net and LSTM models for each environment.  For each group, the decision threshold and corresponding LSTM episode lengths increase from left to right.}
    \label{fig:convergence}
\end{figure}

Figure~\ref{fig:convergence} shows the number of episodes needed for convergence for training the CBGT-Net and LSTM models for each environment.  In all cases, the CBGT-Net required fewer training episodes than the LSTM model.  On average, the CBGT-Net required 75.4\% fewer training episodes than the LSTM model for the MNIST environments, and 89.4\% fewer episodes for the CIFAR-10 environments. In conclusion, the CBGT-Net consistently outperforms the LSTM model in terms of training efficiency across environments.

\section{Conclusion and Future Work}\label{sec:future_work}

This paper introduces a neural network architecture based on cortico-basal ganglia-thalamic circuits found in mammalian brains and demonstrates its effectiveness in learning inference tasks from streams of low-information data.  We demonstrated that the model can learn to categorize images based on a stream of small patches extracted from the image, as well as specify when it should decide based on the amount of supporting evidence observed, as opposed to a fixed number of observations. The model generally outperforms similar models that use LSTMs for recurrent connections and is especially robust to decreasing information presented in each observation. 


In addition to improvements in performance and robustness to low-information observations, the evidence accumulation component provides for transparent deliberation, which we believe offers potential benefits in human-autonomy collaborations.  Specifically, each element in the evidence accumulator corresponds to the model's current preference towards a desired decision, and the margin between accumulator values and decision threshold indicates how imminent a decision may be, as well as the presence of potential alternative decisions that have high levels of accumulated evidence.

There are several potential avenues for future development of the proposed model. Individual components of the model can be extended in multiple ways.  For instance, the accumulator could be extended to incorporate non-linear or temporal dynamics, such as decay, and specifically where dynamics are biologically motivated.
In the current formulation, we asserted that the evidence encoder's dimensionality must match the number of decision categories; future efforts could explore utilizing different representations of evidence and adapting the accumulator accordingly.
One significant direction would be to extend the network to also learn a dynamic decision threshold, as opposed to utilizing a fixed threshold, based on the training environment and/or extrinsic observations ---ensuring training stability and a meaningful interpretation of the threshold would be specific challenges.  The evidence accumulation aspect of the model provides transparency to its deliberation process, providing the opportunity to explore human understanding, interaction, and potential intervention with the model.  Finally, we are interested in applying the model as part of a policy for sequential decision-making tasks, allowing agents to learn to perform actions based on collected evidence, rather than reactively to individual observations.

\bibliographystyle{IEEEtran}
\bibliography{main}

\begin{thebibliography}{10}
\providecommand{\url}[1]{#1}
\csname url@samestyle\endcsname
\providecommand{\newblock}{\relax}
\providecommand{\bibinfo}[2]{#2}
\providecommand{\BIBentrySTDinterwordspacing}{\spaceskip=0pt\relax}
\providecommand{\BIBentryALTinterwordstretchfactor}{4}
\providecommand{\BIBentryALTinterwordspacing}{\spaceskip=\fontdimen2\font plus
\BIBentryALTinterwordstretchfactor\fontdimen3\font minus \fontdimen4\font\relax}
\providecommand{\BIBforeignlanguage}[2]{{%
\expandafter\ifx\csname l@#1\endcsname\relax
\typeout{** WARNING: IEEEtran.bst: No hyphenation pattern has been}%
\typeout{** loaded for the language `#1'. Using the pattern for}%
\typeout{** the default language instead.}%
\else
\language=\csname l@#1\endcsname
\fi
#2}}
\providecommand{\BIBdecl}{\relax}
\BIBdecl

\bibitem{pouyanfar2018-survey}
S.~Pouyanfar, S.~Sadiq, Y.~Yan, H.~Tian, Y.~Tao, M.~P. Reyes, M.-L. Shyu, S.-C. Chen, and S.~S. Iyengar, ``A survey on deep learning: Algorithms, techniques, and applications,'' \emph{ACM Computing Surveys (CSUR)}, vol.~51, no.~5, pp. 1--36, 2018.

\bibitem{su2019-one_pixel}
J.~Su, D.~V. Vargas, and K.~Sakurai, ``One pixel attack for fooling deep neural networks,'' \emph{IEEE Transactions on Evolutionary Computation}, vol.~23, no.~5, pp. 828--841, 2019.

\bibitem{smith2004_decisions}
P.~L. Smith and R.~Ratcliff, ``Psychology and neurobiology of simple decisions,'' \emph{Trends in neurosciences}, vol.~27, no.~3, pp. 161--168, 2004.

\bibitem{maia11_basal_ganglia}
T.~V. Maia and M.~J. Frank, ``From reinforcement learning models of the basal ganglia to the pathophysiology of psychiatric and neurological disorders,'' \emph{Nature Neuroscience}, vol.~14, no.~2, pp. 154--162, February 2011.

\bibitem{stocco2010_basal_ganglia}
A.~Stocco, C.~Lebiere, and J.~R. Anderson, ``Conditional routing of information to the cortex: A model of the basal ganglia’s role in cognitive coordination.'' \emph{Psychological review}, vol. 117, no.~2, p. 541, 2010.

\bibitem{pisauro2017_neural}
M.~A. Pisauro, E.~Fouragnan, C.~Retzler, and M.~G. Philiastides, ``Neural correlates of evidence accumulation during value-based decisions revealed via simultaneous eeg-fmri,'' \emph{Nature communications}, vol.~8, no.~1, pp. 1--9, 2017.

\bibitem{agarwal2018-cbgt}
A.~Agarwal, A.~Kumar~V, K.~Dunovan, E.~Peterson, and T.~Verstynen, ``Better safe than sorry: Evidence accumulation allows for safe reinforcement learning,'' \emph{arXiv preprint arXiv:1809:09147}, 2018.

\bibitem{dunovan2016-believer}
\BIBentryALTinterwordspacing
K.~Dunovan and T.~Verstynen, ``Believer-skeptic meets actor-critic: Rethinking the role of basal ganglia pathways during decision-making and reinforcement learning,'' \emph{Frontiers in Neuroscience}, vol.~10, 2016. [Online]. Available: \url{https://www.frontiersin.org/articles/10.3389/fnins.2016.00106}
\BIBentrySTDinterwordspacing

\bibitem{bond2021-dynamic}
K.~Bond, K.~Dunovan, A.~Porter, J.~E. Rubin, and T.~Verstynen, ``Dynamic decision policy reconfiguration under outcome uncertainty,'' \emph{Elife}, vol.~10, p. e65540, 2021.

\bibitem{moon2020-confidence}
J.~Moon, J.~Kim, Y.~Shin, and S.~Hwang, ``Confidence-aware learning for deep neural networks,'' in \emph{international conference on machine learning}.\hskip 1em plus 0.5em minus 0.4em\relax PMLR, 2020, pp. 7034--7044.

\bibitem{mandelbaum2017-distance}
A.~Mandelbaum and D.~Weinshall, ``Distance-based confidence score for neural network classifiers,'' \emph{arXiv preprint arXiv:1709.09844}, 2017.

\bibitem{lecun2010-mnist}
Y.~LeCun, C.~Cortes, and C.~Burges, ``Mnist handwritten digit database,'' \emph{ATT Labs [Online]. Available: http://yann.lecun.com/exdb/mnist}, vol.~2, 2010.

\bibitem{lecun1998-lenet5}
Y.~LeCun, L.~Bottou, Y.~Bengio, and P.~Haffner, ``Gradient-based learning applied to document recognition,'' \emph{Proceedings of the IEEE}, vol.~86, no.~11, pp. 2278--2324, 1998.

\bibitem{Krizhevsky2009LearningML}
\BIBentryALTinterwordspacing
A.~Krizhevsky, ``Learning multiple layers of features from tiny images,'' 2009. [Online]. Available: \url{https://api.semanticscholar.org/CorpusID:18268744}
\BIBentrySTDinterwordspacing

\bibitem{he2016-resnet}
K.~He, X.~Zhang, S.~Ren, and J.~Sun, ``Deep residual learning for image recognition,'' in \emph{Proceedings of the IEEE conference on computer vision and pattern recognition}, 2016, pp. 770--778.

\bibitem{hochreiter1997-lstm}
S.~Hochreiter and J.~Schmidhuber, ``Long short-term memory,'' \emph{Neural computation}, vol.~9, no.~8, pp. 1735--1780, 1997.

\bibitem{kingma2014-adam}
D.~P. Kingma and J.~Ba, ``Adam: A method for stochastic optimization,'' \emph{arXiv preprint arXiv:1412.6980}, 2014.

\end{thebibliography}

\end{document}